\pdfoutput=1
\PassOptionsToPackage{table}{xcolor}
\documentclass[11pt]{article}

\usepackage[final]{acl}

\usepackage{times,latexsym}
\usepackage{url}

\usepackage{multirow}
\usepackage{adjustbox}
\usepackage{subcaption}

\usepackage{booktabs}
\usepackage{array}
\usepackage{tablefootnote}
\usepackage{amsmath}
\usepackage{scalerel}
\usepackage{siunitx}
\usepackage{tabularx}
\usepackage[table]{xcolor}
\usepackage{soul}
\usepackage{makecell}

\usepackage{inconsolata}
\usepackage{caption}

\usepackage{graphicx}

\usepackage{enumitem}
\usepackage{amsfonts}

\usepackage{dsfont}
\usepackage{amsmath}
\usepackage{amssymb}
\usepackage{bbm}
\usepackage{pifont}

\usepackage{tablefootnote}

\usepackage{color}
\usepackage{comment}

\usepackage{bbding}
\usepackage{algorithm}
\usepackage{algpseudocode}

\usepackage{courier}

\usepackage[T1]{fontenc}

\usepackage[utf8]{inputenc}

\usepackage{microtype}

\usepackage{inconsolata}
\usepackage{authblk}

\title{Multimodal Misinformation Detection by Learning from Synthetic Data with Multimodal LLMs}

\author{
  \textbf{Fengzhu Zeng\textsuperscript{1}},
  \textbf{Wenqian Li\textsuperscript{2}},
  \textbf{Wei Gao\textsuperscript{1}},
  \textbf{Yan Pang\textsuperscript{2}}
\\
  \textsuperscript{1}School of Computing and Information Systems, Singapore Management University \\
  \textsuperscript{2}Institute of Operations Research and Analytics, National University of Singapore
\\
    \href{mailto:fzzeng.2020@phdcs.smu.edu.sg}{fzzeng.2020@phdcs.smu.edu.sg},~
    \href{mailto:wenqian@u.nus.edu}{wenqian@u.nus.edu} \\
    \href{mailto:weigao@smu.edu.sg}{weigao@smu.edu.sg},~
    \href{mailto:jamespang@nus.edu.sg}{jamespang@nus.edu.sg}
}

\begin{document}
\maketitle
\begin{abstract}
Detecting multimodal misinformation, especially in the form of image-text pairs, is crucial.
Obtaining large-scale, high-quality real-world fact-checking datasets for training detectors is costly, leading researchers to use synthetic datasets generated by AI technologies. However, the generalizability of detectors trained on synthetic data to real-world scenarios remains unclear due to the distribution gap. 
To address this, we propose learning from synthetic data for detecting real-world multimodal misinformation through two model-agnostic data selection methods that match synthetic and real-world data distributions. Experiments show that our method enhances the performance of a small MLLM (13B) on real-world fact-checking datasets, enabling it to even surpass GPT-4V~\cite{GPT-4V}.

\end{abstract}

\section{Introduction}
\begin{figure}[t!]
\centering
\includegraphics[width=0.8\linewidth]{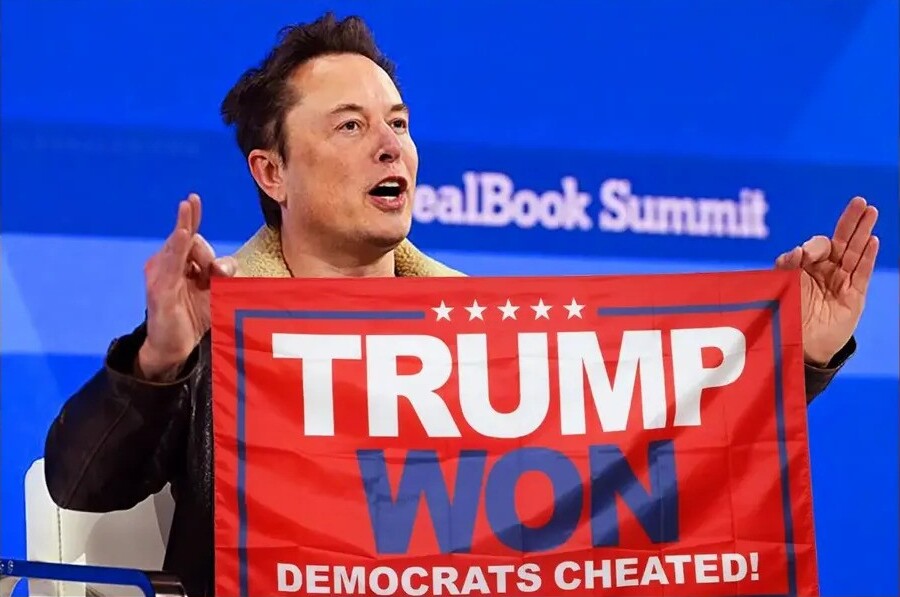}
\caption{Elon Musk holding a flag that says "Trump Won, Democrats Cheated."}
\label{intro:fig}
\end{figure}

Multimodal misinformation, which appears more credible and spreads faster than text-only misinformation, has become a significant concern. About one-third of verification claims include multimodal data, with the primary modality being image-text pairs~\cite{akhtar-etal-2023-multimodal}.
This underscores the importance of Multimodal Misinformation Detection (MMD), which involves determining the veracity of such image-text pairs. These pairs can be created by pairing a textual claim with an out-of-context image, or by manipulating the content of the image, the text, or both. An illustrative example is depicted in Figure~\ref{intro:fig}, where an image shows Elon Musk holding a flag bearing the statement "Trump Won, Democrats Cheated," a false claim debunked by fact-checkers\footnote{\url{https://www.snopes.com/fact-check/elon-musk-trump-won-flag/}}.

The success of training a MMD model highly depends on the availability of large-scale, high-quality datasets, especially in the era of Multimodal Large Language Models (MLLMs) that require instruction tuning for downstream tasks. However, acquiring such real-world fact-checking datasets presents significant challenges as labor-intensive annotation has to be done by fact-checkers. 

Given this data scarcity, a cost-effective solution is to utilize the advancements in generative AI technology to generate synthetic multimodal misinformation based on a vast repository of readily accessible real news~\citep{luo-etal-2021-newsclippings,DGM,jia2023autosplice}. For instance, the synthetic dataset NewsCLIPings~\cite{luo-etal-2021-newsclippings} contains more than 1 million instances, whereas the popular real-world dataset MediaEval~\cite{mediaeval16}, annotated by fact-checkers, only has around 10,000 instances with just 514 unique images. Studies have proposed different multimodal misinformation detectors trained on synthetic datasets and have achieved reasonable performance~\citep{DBLP:conf/cvpr/AbdelnabiHF22,DFC,sniffer}. 

However, it is unclear how well detectors trained on large synthetic datasets can generalize to real-world fact-checking data, as there remain salient gaps between them (Visualization example is shown in Appendix \ref{appendix: vis}).
Since synthetic and real-world data are out-of-distribution (OOD) relative to each other, the distributional gap between the two types of datasets can lead to significant discrepancies in detection performance. 
Detectors directly trained on synthetic datasets may yield limited results or even fail in real-world applications.

To address this gap, we propose to select valuable synthetic data instances based on similarity metrics that are oblivious to the downstream detection tasks, and utilize it to improve the generalization capability of detectors to real-world fact-checking data.
We first compile a large-scale synthetic dataset by amalgamating three distinct synthetic datasets, ensuring coverage of diverse multimodal misinformation categories. 
Given this synthetic dataset as training set and \textit{a very small number of unlabeled} real-world instances as the validation set, our goal is to effectively select \textit{a small subset} of relevant and valuable synthetic data to enhance the models' capacity for detecting real-world multimodal misinformation.

We approach this using two model-agnostic data selection methods: 1) semantic similarity-based selection, which prioritizes synthetic instances with the highest similarity scores to the validation set; and 2) distributional similarity-based selection using gradient information derived from the Optimal Transport (OT) problem~\cite{villani2009optimal} to increase the density of data around the desirable region by choosing data points in the synthetic dataset that are close to the target real-world distribution.
These methods evaluate the relevance of synthetic data points without requiring model retraining, ensuring efficiency and scalability. Moreover, the small selected subset of synthetic training instances makes the fine-tuning of MLLMs feasible in computing resource-constrained scenarios. 

Our main contributions are four-fold~\footnote{Code and dataset are released at \url{https://github.com/znhy1024/MLLMs-for-MMD}.}:
\begin{itemize}[leftmargin=*]

    \item We propose a new task to tackle the scarcity of real-world fact-checking data for MMD by learning from synthetic data. 

     \item We frame a new setting for MMD in the era of MLLMs: how can we effectively select relevant and valuable instances from a large-scale synthetic dataset to improve MLLM's detection of real-world multimodal misinformation?

    \item We employ two model-agnostic data selection methods to handle different distributions by selecting a small number of synthetic instances for fine-tuning open-source MLLMs on the MMD task. 
 
    \item Evaluation on real-world fact-checking datasets demonstrates the effectiveness of data selection methods across MLLM scales and families. This approach enables a small MLLM (13B) to even surpass GPT-4V~\cite{GPT-4V}.
\end{itemize}

\section{Problem Formulation} 
Let $\mathcal{D}= \{(x_i,y_i)\}_{i=1}^{|\mathcal{D}|}$ be a multimodal misinformation dataset with $|\mathcal{D}|$ instances, where each instance consists of the input $x_i$ and the ground-truth label $y_i \in \{0,1\}$. We denote $x_i=(\mathbf{m}_i,\mathbf{t}_i)$ which is a pair of information containing an image $\mathbf{m}_i$ and a corresponding claim text $\mathbf{t}_i$, and $y_i$ indicates the veracity of $x_i$ is true if $y_i=0$ or false otherwise. 
The task of MMD aims to determine the veracity label of the given piece of information $x_i$. 

We consider a large dataset $\mathcal{D}_s$ consisting of synthetic instances that cover common categories of multimodal misinformation, and a target dataset $\mathcal{D}_t$ which contains high-quality, annotated, and diverse real-world fact-checking data, where $|\mathcal{D}_s| \gg |\mathcal{D}_t|$.  
Our goal is to select a subset $\mathcal{D}_v=\{(x_i,y_i)\}_{i=1}^{|\mathcal{D}_v|}$ from $\mathcal{D}_s$ to fine-tune a pre-trained multimodal model $\mathcal{M}$, such that it generalizes well to the real-world test dataset $\mathcal{D}_t$.  
To reflect real-world constraints related to practical computational resource limitations, we also restrict the size of the selected subset being small, i.e. $|\mathcal{D}_v| \ll |\mathcal{D}_s|$.

A \textit{small unlabeled} validation set $\mathcal{D}_u=\{(x_i)\}_{i=1}^{|\mathcal{D}_u|}$ with a few randomly sampled instances from $\mathcal{D}_t$ is given to facilitate data selection. We assume the unlabeled validation set is sampled i.i.d from the target distribution as they originate from the same source. It is expected that MLLMs trained on the selected synthetic instances based on $\mathcal{D}_u$ could generalize well on unseen test instances.

\section{Multimodal Misinformation Data} \label{data intro}

\subsection{Common Categories} 
There is no standard categorizations for multimodal misinformation. We categorize it into two types based on whether the image content are falsely altered or counterfeited, as briefly described below. Appendix \ref{sec:appendix-dataset-cc} contains a more formulated and detailed description.

\noindent\textbf{Out-of-Context (OOC) Misuse} occurs when the textual claim misrepresents the original context or intent of a genuine image, conveying misleading information. OOC image-text pair can be obtained by pairing a textual claim with an image taken out-of-context~\citep{MIAM,luo-etal-2021-newsclippings}, or manipulating the textual claim such as replacing named entities and altering sentiments, aiming at distorting its meaning to conform to a false narrative~\citep{MEIR,TempNews}. Such OOC pairs not only deceive human but also pose significant challenges for detection models~\cite{luo-etal-2021-newsclippings}.

\noindent\textbf{Content Manipulation} involves intentionally altering image and text modality with the aim of deceiving or misleading. As it can be difficult to consistently obtain real images to support non-factual claim, manipulating image and text content to generate misinformation becomes an alternative approach~\cite{mm-mis-detect}. Image manipulation refers to altering an image to obtain its fake version by modifying the elements within it using techniques, such as removing content and face swap~\citep{DGM,jia2023autosplice}. Image and text manipulation techniques can be combined to ensure that both visual and textual elements complement each other, thereby strengthening the deception.

\subsection{Training and Evaluation Datasets} \label{training data}
Given that synthetic datasets often lack the comprehensive diversity of misinformation categories found in real-world data and typically focus on specific types, such as out-of-context misuse, we curate a large-scale synthetic training dataset by including three representative synthetic datasets. This ensures coverage of common multimodal misinformation categories and includes diverse techniques for creating more deceptive multimodal misinformation.

For a comprehensive assessment of model's performance across various sources and distributions of misinformation, we utilize two real-world fact-checking datasets collected from social media and fact-checking websites, both of which provide sufficient data for reliable evaluation. 
Appendix \ref{sec:appendix-dataset-te} contains more datasets details.

\textbf{Training Datasets}: 1) \underline{NewsCLIPpings}~\cite{luo-etal-2021-newsclippings} is an automatically generated OOC multi-modal dataset that contains both pristine and falsified instances. A falsified instance is an unmanipulated but mismatched image-caption pair, constructed by pairing an image with a caption from an inconsistent context. The falsified instances could portray subjects in an image as different entities, depict specific individuals in a misleading context, and mislabel the event described within a particular scene. By utilizing different embeddings models during the image-text mismatch procedure, around 988k unique instances are automatically generated.
2) $\underline{\text{DGM}^4}$~\cite{DGM} is a multimodal misinformation dataset, where synthetic instances are automatically generated by applying image and text manipulation techniques to pristine instances sourced from news data. By employing diverse manipulation techniques (e.g., face swap, emotion manipulation, name entity replacement), eight manipulated instances comprising both OOC and manipulation instances are generated for each pristine instance, resulting in a total of 230k synthetic instances.
3) \underline{Autosplice}~\cite{jia2023autosplice} is a manipulated image dataset that utilizes a language-image model DALLE-2~\cite{delle2}, guided by textual prompts to edit images. Unlike DGM$^4$, where image and text manipulation are initially employed separately and then combined, Autosplice alters the image based on modified text, enabling a more direct and integrated manipulation process.

\textbf{Evaluation Datasets}: 1) \underline{MediaEval}~\cite{mediaeval16} collects a set of tweets associated images/videos with manually verified veracity (i.e., fake or real), which were spread around 17 widely attention-grabbing events. The number of instances for each event is small, and the classes are imbalanced. We utilize the image-text instances for evaluation, consisting of 702 instances, in which 292 are real and 410 are fake. 2) \underline{Snopes}, a dataset derived from Fauxtography~\cite{zlatkova-etal-2019-fact} and MOCHEG~\cite{end2end-mm}, which contains image-related claims from the popular fact-checking website Snopes\footnote{\url{Snopes.com}}. These image-claim pairs are labeled as either True or False by fact-checkers, consisting of 756 instances, where 376 are true and 380 are false.

\section{Methodology}
\label{dataselection}

Data selection aims to find a set of training instances that improve model performance~\cite{DBLP:journals/corr/abs-2402-16827}. Some approaches use data valuation and attribution methods such as influence function~\cite{hampel1974influence,koh2017understanding,xia2024less}, Leave-One-Out~\cite{hastie2009elements}, and Shapley value~\cite{ghorbani2019data}. These approaches perturb one data point from the training set to get a new subset, with or without permutations, to trace the change of the model validation performance, and quantify the effect of that point. However, such process requires re-training the model on each subset that excludes each perturbed data point to be evaluated, which will cause expensive computational costs~\cite{TRAK}. Additionally, it requires labeled validation sets to trace performance changes, which differs from our setting.

Recently, distributional distance metrics such as Wasserstein distance and KL divergence, have emerged as effective proxies for conducting data selections by providing an upper bound on performance change~\citep{courty2017joint,nguyen2021kl,just2023lava,li2023data,li2024private}. As model-agnostic approaches, they are more desirable in the context of LLMs. There have been some successful applications for either pre-training or pre-fine-tuning~\citep{XieS0L23,everaert2023gio,kang2024get}. For example, DSIR~\cite{XieS0L23} proposes importance resampling on the hashed n-gram features of text data, and measure the distributional similarity based on KL-reduction.
GIO~\cite{everaert2023gio} uses iterative gradient methods to prune training samples by minimizing KL-divergence. GOT-D~\cite{kang2024get} enjoys the geometric property of Optimal Transport to select data that nudges the pre-training distribution closer to the target distribution. However, these methods are not directly applicable to multimodal data due to the difficulties in quantifying the probability density of multimodal misinformation data, typically necessitating tens of thousands of data points. Additionally, applying these textual data selection methods in the context of MMD task is incorrect. The factuality of a piece of information is determined by the combined information from text and image in MMD. Relying solely on a single modality fails to capture cases like out-of-context misuse, where the standalone textual claim might be accurate while the image-text pair together conveys false information.  

Several studies in multimodal data selection primarily focus on sampling large corpora, ranging from millions to billions of samples, from vast volumes of noisy, web-curated datasets~\citep{DBLP:conf/cvpr/RaoRNDCD20,DataComp,DBLP:journals/corr/abs-2309-15954,T-MARS,DBLP:journals/corr/abs-2403-02677}. These methods involve non-trivial data-dependent filtering strategies to select high-quality data for pre-training, such as CLIPScore~\cite{hessel-etal-2021-clipscore} that ranks each data point based on the cosine similarity scores between its CLIP image and text embeddings. 
Different from these works, our goal is to select a small number of useful synthetic instances for fine-tuning, with the aim of aligning distributions between synthetic and real-world datasets for MMD. Another line of work focuses on selecting features or modalities to boost model performance~\citep{KAMYAB2016586,10.1145/3343031.3350987,Lu_2021_ICCV,10132437,JMLR:v25:23-0439}, which is beyond the scope of this paper.

We employ two model-agnostic data selection approaches, i.e., semantic similarity-based selection and distributional similarity based selection, which provide straightforward, efficient, and scalable estimates of synthetic data relevance, facilitating the tuning of MLLMs for detecting real-world multimodal misinformation. 

\paragraph{Feature Extraction.} We adopt a mixed-modal encoder, consisting of a text encoder and an image encoder, to extract multimodal features for data selection. It is a simple extension of the off-the-shelf CLIP model~\cite{CLIP}, which has demonstrated good performance in multimodal retrieval tasks~\cite{racm3}. Specifically, given an image-claim pair $x_i$, we obtain the visual features and textual features via the image encoder and text encoder. This pair's multimodal features $\mathbf{e}_i$ are then computed by averaging these two modalities, with the $L_2$ norm scaled to $1$.

\paragraph{Semantic Similarity~(SemSim).} 
Similarity between the synthetic training set $\mathcal{D}_s$ and the validation set $\mathcal{D}_u$ can be used for data selection. Specifically, the semantic features of instances in each set are obtained using the mixed-modal encoder. Then, we calculate the similarity between the multimodal features $\mathbf{e}_i$ of a training instance and the averaged multimodal features $\mathcal{E}_u=\frac{1}{|\mathcal{D}_u|}\sum_{j=1}\mathbf{e}_j$ of the validation set $\mathcal{D}_u$. Subsequently, we select the data points with the highest similarity score to construct $\mathcal{D}_v$. We use the cosine similarity as the measure, defined as follows:
\begin{equation}
    \mathcal{G}_{\text{ss}}(\mathbf{e}_i, \mathcal{E}_u) = \frac{\mathbf{e}_i\cdot \mathcal{E}_u}{||\mathbf{e}_i||\cdot ||\mathcal{E}_u||}.
\end{equation}

\paragraph{Distributional Similarity (DisSim).} 
Wasserstein distance~\cite{kolouri2017optimal} is a distributional measure and has been shown to provide an upper bound on the difference in a model’s performance when it is trained on one distribution and evaluated on another. With this in mind, the selection strategy for the featurized synthetic set $\mathbf{e}^s$ is to prioritize
data points that could increase the density around the featurized target set $\mathbf{e}^u$.
Formally, the $p$-Wasserstein distance between two probability measures $\mathbf{e}^s$ and $\mathbf{e}^u$ is defined as follows:
\begin{align}
\label{primal}
&\mathcal{W}_p(\mathbf{e}^s,\mathbf{e}^u) \nonumber\\
\stackrel{(a)}{=}&\Big(\inf_{\pi \in \Pi(\mathbf{e}^s,\mathbf{e}^u))} \int c^p(\mathbf{e}_i^s,\mathbf{e}_j^u)d\pi(\mathbf{e}_i^s,\mathbf{e}_j^u)\Big)^{\frac{1}{p}},\nonumber\\
\stackrel{(b)}{=}&\max_{(f,g)\in C^0(\mathcal{X})^2}\langle~f, \mathbf{e}^s~\rangle+\langle~g, \mathbf{e}^u~\rangle 
\end{align}
where the equality $(a)$ comes from the definition of the Wasserstein distance, which is the primal problem. Specifically, $c^p(\mathbf{e}_i^s,\mathbf{e}_j^u)=||\mathbf{e}^s_i-\mathbf{e}^u_j||^p_p$ and $p\geq 1$ for $i\in\{1,\cdots,|\mathcal{D}_s|\}$ and $j\in \{1,\cdots,|\mathcal{D}_u|\}$, represents the pairwise distance metric.  Without loss of generality, we set $p=2$.
$\pi\in\Pi(\mathbf{e}^s,\mathbf{e}^u)$ is the joint distribution of $\mathbf{e}^s$ and $\mathbf{e}^u$, and any $\pi^\star$ attains the minimum value of equality $(a)$ is considered as an OT plan~\cite{villani2009optimal}, as it is the most cost-effective strategy to make synthetic set be transformed into the target set. Kantorovich formulation~\cite{kantorovich1942translocation} defines the OT problem as a Linear Program. Then, based on the duality theorem,  
the equality $(b)$ holds if $\pi^\star$ and $(f^\star,g^\star)$, are optimal variables of the corresponding primal and dual problem, respectively. Specifically, $f\in\mathbb{R}^{|\mathcal{D}_s|}$ and $g\in \mathbb{R}^{|\mathcal{D}_u|}$, and $C^{0}(\mathcal{X})$ is the set of
all continuous functions over the feature space $\mathcal{X}$.

We select the highest-scored data points with the largest negative calibrated $\emph{wasserstein gradient}$, defined as follows:
\begin{equation}
\mathcal{J}_{\text{wg}}(\mathbf{e}^s_i,\mathbf{e}^u)= f_i^\star - \sum_{{j\in \{1,\cdots,|\mathcal{D}_s|\}\backslash i }}\frac{f_j^\star}{|\mathbf{e}^s|-1},
\end{equation}
which measures the sensitivity of the $i$-th data point of the synthetic training set for $\mathcal{W}_p(\mathbf{e}^s,\mathbf{e}^u)$. This gradient value determines the shifting direction based on whether it is positive or negative. If the value is positive (negative), shifting more probability mass to that datum will result in an increase (decrease) of the distance between the synthetic set and the validation set.

\begin{table*}[t!]
\small
\centering
\begin{adjustbox}{width={\linewidth},keepaspectratio}%
\begin{tabular}{l|c|cc|c|ccc}
\toprule[1.0pt]
  \multicolumn{2}{c}{\textbf{Setting}}
&\multicolumn{2}{|c}{Direct Prompting} & \multicolumn{1}{|c}{Real-world}& \multicolumn{3}{|c}{Synthetic}\\
\cmidrule{1-8} 
      \textbf{Method}& Majority & LLaVA &GPT-4V & $\text{LLaVA}_{\text{M}\rightleftharpoons\text{S}}$   & $\text{LLaVA}_{\text{R}}$      & $\text{LLaVA}_{\text{S}}$   & $\text{LLaVA}_{\text{D}}$    \\ 
\midrule[0.5pt]
{MediaEval}
                      & ${0.368}$  & ${0.480}$ & $0.595$& ${0.431}$  &${0.568}_{(0.020)}$ &$\mathbf{0.687}_{(0.015)}$ &$\underline{0.611}_{(0.009)}$ \\                
\midrule[0.5pt]
$\text{Snopes}$
                   & ${0.335}$  & ${0.407}$ & $\mathbf{0.614}$& ${0.511}$  &$\underline{0.540}_{(0.020)}$ &${0.521}_{(0.022)}$ &${0.496}_{(0.027)}$  \\
\specialrule{0em}{1pt}{1pt}              
Snopes (O+)
                   & ${0.335}$  & ${0.399}$ & $0.394$& ${0.548}$  &$\underline{0.758}_{(0.024)}$ &${0.716}_{(0.037)}$ &$\mathbf{0.813}_{(0.017)}$  \\

\bottomrule[1.0pt]
\end{tabular}
\end{adjustbox}
\caption{Results of Majority, direct prompting LLaVA and GPT-4V, full-dataset fine-tuning on the other real-world dataset~($\text{LLaVA}_{\text{M}\rightleftharpoons\text{S}}$), and fine-tuning on a small set of selected synthetic data~($\text{LLaVA}_{\text{R}}$: Random, $\text{LLaVA}_{\text{S}}$: SemSim, $\text{LLaVA}_{\text{D}}$: DisSim). We report the macro-F1 averaged over 3 trials with different random seeds. The best results for each dataset are in \textbf{bold} while the second-best results are \underline{underlined}. The standard deviation is in (.).}
\label{tab: main results}
\end{table*}

\section{Experimental Evaluation}
\subsection{Experimental Settings}
\paragraph{Datasets.} As introduced in \S \ref{data intro}, we uses NewsCLIPpings~\cite{luo-etal-2021-newsclippings}, $\text{DGM}^4$~\cite{DGM} and Autosplice~\cite{jia2023autosplice} to construct a large synthetic dataset as training set. We evaluate models on the \textbf{MediaEval}~\cite{mediaeval16} and the \textbf{Snopes} datasets. 
Since the Snopes dataset only includes instances labeled as True/False but excludes those labeled as Miscaptioned\footnote{\url{https://www.snopes.com/fact-check/rating/miscaptioned/}}~\cite{zlatkova-etal-2019-fact} that fall into OOC category, we observe a disproportionately high volume of manipulation instances.  
For a balanced evaluation, we additionally create a variant dataset \textbf{Snopes (O+)} with increased OOC ratio based on the original one by mismatching the image and text of around 50\% instances of Snopes.

\paragraph{Base MLLMs.} We employ LLaVA-NeXT-13B\footnote{In the rest of the paper, we refer to LLaVA-NeXT-13B as LLaVA for brevity.}~\cite{liu2024llavanext} as the base MLLM for fine-tuning, given its excellent performance on various multimodal tasks compared to other MLLMs. Additionally, we perform an ablation study on different base models, including LLaVA-7B and mPLUG-Owl2~\cite{mPLUG-Owl2}. More details are in Appendix \ref{sec:appendix-exp-mllm}.

\paragraph{Baselines.} We compare with full-dataset fine-tuned LLaVA~\cite{liu2024llavanext} models: $\text{LLaVA}_{\text{M}\rightleftharpoons\text{S}}$, which are the models 1) trained on full set of MediaEval and evaluated on Snopes and Snopes (O+); and 2) trained on Snopes and evaluated on MediaEval. 
We also compare with the random selection, where we randomly sample instances from the synthetic dataset for fine-tuning. We also include baselines using strong MLLMs as misinformation detectors, including LLaVA~\cite{liu2024llavanext} and GPT-4V~\cite{GPT-4V}, where we directly prompt them to generate predictions.

\paragraph{Default Settings.} We utilize the off-the-shelf CLIP model (ViT-L/14)~\cite{ViT-L/14} for feature extraction. We use the same prompt template for all models to ensure a fair comparison. 
Each data selection method empirically selects $750$ synthetic instances from the synthetic dataset $\mathcal{D}_s$ to construct $\mathcal{D}_v$ for fine-tuning, and we ensure the class distribution of selected set is balanced. The unlabeled validation set $\mathcal{D}_u$ contains 5\% instances randomly sampled from each test set.
We conduct all experiments with the selection methods three times using different random seeds, and report the mean macro-F1 and standard deviation of each metric across these three runs. More details are provided in Appendix \ref{sec:appendix-exp-default}.

\begin{figure*}[t]
  \centering
  \begin{subfigure}[b]{0.33\linewidth}
    \centering
    \includegraphics[width=\linewidth]{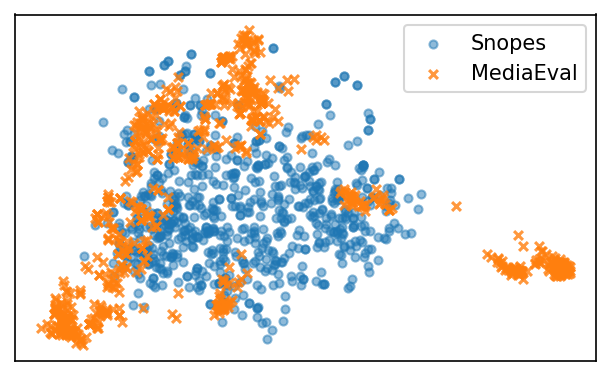}
    \caption{}
    \label{fig:testdata1}
  \end{subfigure}%
  \begin{subfigure}[b]{0.33\linewidth}
    \centering
    \includegraphics[width=\linewidth]{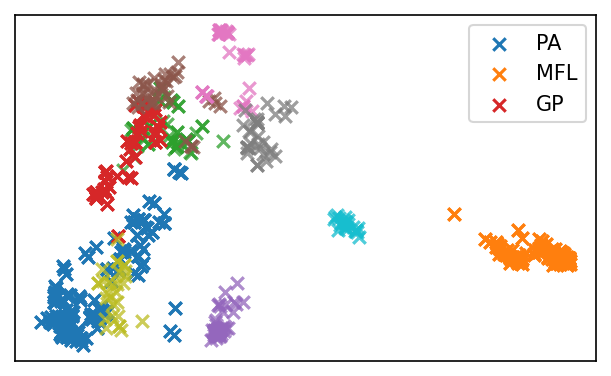}
    \caption{}
    \label{fig:testdata2}
  \end{subfigure}
  \begin{subtable}[b]{0.33\linewidth}
    \centering
    \small
    \begin{tabular}{l|ccc}
    \toprule[1.0pt]
     Model & PA  & MFL   & GP    \\
    \midrule[1.0pt] 
    SemSim & \makecell{0.781 \\ \tiny{(0.020)}} & \makecell{0.356 \\ \tiny{(0.115)}}  &\makecell{0.323 \\ \tiny{(0.078)}}\\  
    \midrule[0.5pt] 
    DisSim &\makecell{0.822 \\ \tiny{(0.027)}} & \makecell{0.333 \\ \tiny{(0.077)}}  &\makecell{0.375 \\ \tiny{(0.044)}}\\  
    \bottomrule[1.0pt]
    \end{tabular}
    \vspace{0.11cm}
    \caption{}
    \label{fig:testdata3}
  \end{subtable}
  \caption{The 2D projection of the multimodal features using PCA. \textbf{(a)} MediaEval and Snopes datasets; \textbf{(b)} Top-10 events with the most instances in MediaEval. Each colored group represents an event, and three highlighted groups are top-3 events; \textbf{(c)} The F1 score of semantic and distributional selection methods on top-3 events with the most instances from MediaEval dataset. (.) encloses standard deviation. PA: Paris Attack; MFL: Mt Fuji Lenticular; GP: German Protest.}
  \label{fig:testdata_combined}
\end{figure*}

\subsection{Experimental Results}

\subsubsection{Main results and analysis}

We present main results in Table \ref{tab: main results}, and provide an in-depth analysis.

\paragraph{Directly prompt base model.} In Table~\ref{tab: main results}, we observe that directly prompting the LLaVA model does not yield satisfactory performance, indicating that merely relying on the inherent knowledge of base MLLM is insufficient for MMD, especially when the model size is not large enough. We conjecture that task-specific information might be necessary to induce this capability in the base model.  

\paragraph{Utilize OOD real-world data.}
However, incorporating task-specific knowledge by training on one real-world fact-checking dataset does not consistently enhance performance on another; it may even impede it. As shown in Table \ref{tab: main results}, training LLaVA on the MediaEval dataset enhances its detection performance on Snopes and Snopes (O+) with an absolute increase of $0.104$ and $0.149$ respectively, but the reverse adversely affects its performance, where we observe a decline of $0.09$. We examine this inconsistency by checking the distributions of Snopes and MediaEval datasets. 

Specifically, we employ Principal Component Analysis (PCA)~\cite{PCA} to project the multimodal features of each instance from both two datasets. As depicted in Figure~\ref{fig:testdata1}, the MediaEval dataset exhibits a feature distribution characterized by  multiple clusters, displaying OOD traits when compared across different clusters. This is attributed to MediaEval encompassing various events, each comprising posts with high internal density. This observation is further supported by the visualization of the top-10 events with the most instances in MediaEval in Figure~\ref{fig:testdata2}.

In contrast, the distribution of the Snopes dataset lacks such characteristics, as the instances comprise independent claims from fact-checking websites.
Therefore, training on such diverse yet insufficient instances might induce catastrophic forgetting behaviors, leading to dramatic performance degradation. Conversely, when there are more similar instances within each cluster of MediaEval, the model can learn the general semantic patterns from each group more effectively, thereby enhancing its capability in misinformation detection.
These findings suggest that direct training on real-world data cannot guarantee robust generalization, particularly when these real-world data are often OOD relative to each other.

\begin{table*}[t!]
\small
\centering
\begin{adjustbox}{width={0.9\linewidth},keepaspectratio}%
    \begin{tabular}{lccccccccc}
    \toprule[1.0pt]
      & \multicolumn{4}{c}{\textbf{LLaVA-7B}} & & \multicolumn{4}{c}{\textbf{mPLUG-Owl2}}    \\
      \cline{2-5} \cline{7-10}
        \specialrule{0em}{1pt}{1pt}
        & Prompt &  SemSim  &DisSim &$\Delta$  &  &  Prompt &  SemSim  & DisSim &$\Delta$ \\ 
    \midrule[1.0pt]
    MediaEval &0.410  & \makecell{ \textbf{0.667} \\ \tiny{(0.019)}}  &\makecell{ 0.621 \\ \tiny{(0.057)}} &\cellcolor{green!20}0.257$\uparrow$
    & &0.294  & \makecell{\textbf{0.685} \\ \tiny{(0.069)}}  &\makecell{{0.412} \\ \tiny{(0.037)}} &\cellcolor{green!20}0.391$\uparrow$\\  
    \midrule[0.5pt] 
    Snopes & 0.337 & \makecell{  \textbf{0.489} \\ \tiny{(0.033)}}  &\makecell{{0.462} \\ \tiny{(0.034)}}&\cellcolor{green!20}0.152$\uparrow$
    & & 0.332 & \makecell{0.368 \\ \tiny{(0.036)}}  &\makecell{\textbf{0.405} \\ \tiny{(0.058)}} &\cellcolor{green!20} 0.073$\uparrow$\\  
    \specialrule{0em}{1pt}{1pt}              
\text{Snopes (O+)} & 0.373 & \makecell{  0.689 \\ \tiny{(0.009)}}  &\makecell{\textbf{0.752} \\ \tiny{(0.042)}}&\cellcolor{green!20}0.379$\uparrow$
    & & 0.285 & \makecell{0.386 \\ \tiny{(0.014)}}  &\makecell{\textbf{0.402} \\ \tiny{(0.037)}} &\cellcolor{green!20} 0.117$\uparrow$\\
    \bottomrule[1.0pt]
    \end{tabular}
    \end{adjustbox} 
    \caption{Results of MLLMs with different model scales and families. (.) encloses standard deviation.}
\label{tab: base model ablation}
\end{table*}

\begin{figure*}[t]
  \centering
  \includegraphics[width=0.9\linewidth]{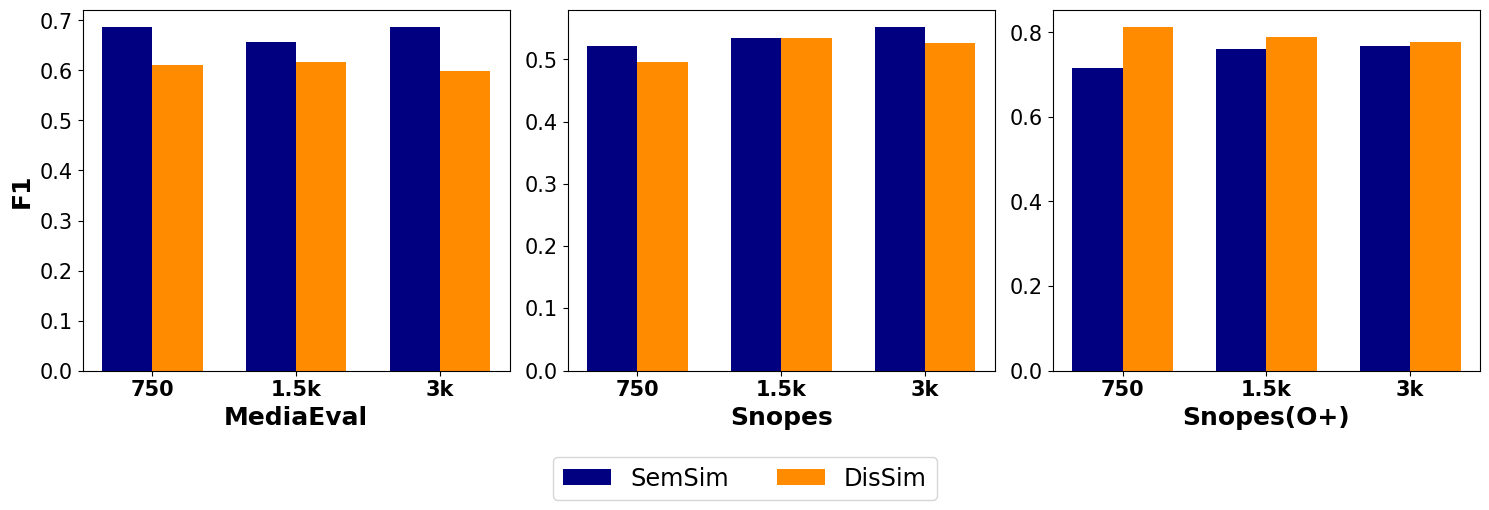}
  \caption{The F1 score with increasing the number of selected synthetic instances for training.}
  \label{fig: training size}
\end{figure*}

\paragraph{Utilize synthetic data.}
Compared to using OOD real-world data, a small amount of selected synthetic data consistently enhances base model's performance on various real-world datasets, with absolute F1 score improvements ranging from $0.133$ to $0.414$, as observed in Table \ref{tab: main results}. This highlights the promising potential of leveraging synthetic data for real-world MMD. 
Next, we provide a detailed analysis of specific results from the standpoint of data distribution, offering insights for selecting synthetic data for real-world MMD.

Firstly, both similarity-based data selection methods, SemSim and DisSim, enable the base MLLM to outperform GPT-4V on MediaEval and Snopes (O+) datasets respectively. This confirms the effectiveness of these methods and demonstrate their applicability in real-world scenarios. In Table \ref{tab: main results}, SemSim achieves the best F1 score on the MediaEval dataset ($0.687$) surpassing DisSim ($0.611$), while DisSim exhibits the best on the Snopes (O+) dataset ($0.813$) outperforming SemSim ($0.716$).
We hypothesize this phenomenon is influenced by the choice of similarity metrics.

Semantic similarity and Wasserstein distance, provide distinct criteria for evaluating the similarities of data points. Semantic similarity primarily assesses the resemblance of individual points in feature space, selecting data samples that closely match the representations of the target set. In contrast, Wasserstein distance excels at discerning similarities across different distributions, prioritizing the construction of the mean distribution among all probability measures to minimize the total transport cost, rather than matching individual clusters. To delve deeper, we perform data selection and evaluation for each of the top-3 events (as other events lack sufficient instances) and report the macro-F1 in Table~\ref{fig:testdata3}. We find that when selecting synthetic data for a specific event, DisSim outperforms SemSim by $5\%$ on average, aligning with our earlier hypothesis. 
These insights suggest the importance of employing appropriate criteria for data selection to enhance results on real-world data.

Secondly, all the data selection methods improve the base model on Snopes dataset, narrowing the performance gap between the base MLLM and GPT-4V. However, the magnitude of the improvements become smaller compared to the MediaEval and Snopes (O+) datasets. For example, as shown in Table \ref{tab: main results}, DisSim improves the base model performance from $0.399$ to $0.813$ on Snopes (O+), whereas its improvement is from $0.407$ to $0.496$ on Snopes. The reason is likely two-fold: 1) These two datasets have different evaluation focuses--Snopes requires models to have strong manipulation detection capabilities while Snopes (O+) emphasizes OOC detection. Additionally, OOC instances are easier to obtain than manipulation instances requiring data-specific adjustments, which may result in more OOC instances in the synthetic data. Consequently, when the synthetic data distribution has more OOC instances as support, DisSim can better identify valuable instances based on gradient direction and magnitude, aligning the training distribution of the base MLLM closer to the target distribution. 2) Detecting image manipulation usually requires the base MLLMs have strong visual grounding capabilities, while most open-source MLLMs exhibit some systematic visual shortcomings because their pre-trained CLIP vision encoders might overlook visual details in images~\cite{Eyes-Wide-Shut}.
These observations suggest the importance of increasing diversity of misinformation data in synthetic datasets and enhancing the capability of vision encoder for better real-world MMD.

\subsubsection{Impact of base MLLMs}
In Table \ref{tab: base model ablation}, we report the results of SemSim and DisSim across different base models. Fine-tuning on a small number of synthetic data selected by both methods consistently improves the performance of the base models, including MLLMs with smaller model size (i.e., LLaVA-7B) and from different families (i.e., mPLUG-Owl2~\cite{mPLUG-Owl2}). The maximum F1 score improvement on all datasets for LLaVA-7B is $0.26$ on average and for mPLUG-Owl2 is $0.19$, which indicates that the selected synthetic data is generalizably useful, enhancing the MMD performance of various MLLMs. Notably, as SemSim and DisSim are model-agnostic data selection approaches, the selected synthetic data can be reused without further selection costs, endowing efficiency to their deployment on real-world data.

\subsubsection{Impact of the number of selected data}
We investigate whether increasing the number of selected synthetic instances improves the performance of the base MLLM. The overall results in Figure \ref{fig: training size} indicate that increasing the number of selected synthetic instances does not necessarily improve the performance of the base MLLM. A small number of selected synthetic data already contain data points with most of the relevant information for MMD on real-world data. We observe that increasing selected training instances slightly hinders its performance on MediaEval and Snopes (O+). This is because, on MediaEval, as the size increases, the selected data instances tend to converge towards a ``mean'' distribution of all events as discussed before, thus deviating further from the distribution of individual cluster. And we hypothesize that the small number of selected data by DisSim on Snopes (O+) already encapsulates the traits of OOC misinformation, and more data may introduce additional irrelevant or detrimental data points. Determining an optimal number is a problem beyond the scope of our current work.

\section{Related Work} 

\paragraph{Multimodal Misinformation Datasets.}

\underline{Real-world datasets} are typically collected from social media posts like Twitter~\citep{Faking-Sandy,mediaeval16}, or fact-checking websites like Snopes~\citep{zlatkova-etal-2019-fact,end2end-mm}, but such collection processes are labor-intensive and costly. Another line of datasets consist of images and captions collected from Reddit\footnote{\url{https://www.reddit.com/}} posts, with each instance labeled based on subreddit themes~\citep{PS-Battles,In-the-Wild,nakamura-etal-2020-fakeddit}. For example, all posts from the subreddit "usnews" are labeled as real since they are authentic news, while the posts from the subreddit "photoshopbattles" are labeled as fake since this subreddit is for users to battle using image manipulation software. However, these datasets do not reflect the complexities of real-world circulating misinformation. 

Recently, \underline{synthetic datasets} have emerged as a cost-effective data solution~\citep{COSMOS,luo-etal-2021-newsclippings,DGM,jia2023autosplice}, which can be generated in a large scale by using generative AI technology. For instance, the synthetic datasets NewsCLIPings~\cite{luo-etal-2021-newsclippings} swap the captions of different news images, and MEIR manipulates the news captions~\citep{MEIR} to create OOC misinformation. On the other hand, some datasets manipulate the image of news and combine it with text manipulation to generate fake versions~\citep{DGM,jia2023autosplice}. 
However, such datasets might not reflect the data distribution of real-world misinformation.
In this paper, we aim at exploring how well the models trained on synthetic data can generalize to real-world fact-checking datasets.

\paragraph{Multimodal Misinformation Detection.}
Most existing methods for MMD are typically trained in a fully supervised manner using the entire training set and evaluated on the test set from the same small-scale real-world fact-checking dataset~\citep{DBLP:conf/mm/JinCGZL17,EANN,MVAE,wu-etal-2021-multimodal,HMCAN,FNR,MMFN,liu-etal-2023-interpretable,chen-etal-2023-causal}, or from the same large-scale synthetic dataset~\citep{COSMOS,DBLP:conf/cvpr/AbdelnabiHF22,DGM,jia2023autosplice,Mu_2023_WACV,DFC,sniffer}. 
Meanwhile, by integrating external world knowledge, \citet{LEMMA} improve the performance of GPT-4V~\cite{GPT-4V} on real-world fact-checking datasets. \citet{FakeNewsGPT4} improve cross-domain performance of PandaGPT~\cite{PandaGPT} on a synthetic dataset, which allows the model trained on instances generated using news from one agency (e.g., BBC) to generalize to those generated using news from other sources (e.g., USA TODAY).
Given the discrepancy in size between synthetic and real-world datasets, our work aims to enhance the generalizability of models trained on synthetic data to test on real-world misinformation datasets using data selection methods, which provides an cost-effective solution for MMD.

\section{Conclusion and Future Work}
We propose leveraging synthetic data to address the MMD task in the scenarios of real-world data scarcity, in which relevant and valuable instances can be selected from a large-scale synthetic dataset for fine-tuning MLLMs. We advocate for utilizing two model-agnostic similarity-based data selection approaches for this task.
Results demonstrate that even using a very small number of selected synthetic training instances can significantly boost MLLMs' detection performance on real-world fact-checking data, enabling a small MLLM, LLaVA-13B, to outperform GPT-4V. 

In future research, we plan to determine the optimal number of selected instances for different data selection methods for maximizing the effectiveness of useful training data. Additionally, we plan to explore more design solutions to obtain better multimodal features for further improving misinformation detection performance. 

\section{Limitations}
Although a small number of selected synthetic data can significantly boost the base MLLM's detection performance, there might be an optimal number of synthetic instances that yield best results. Therefore, determining such optimal numbers is crucial to optimize the effectiveness of training instances. 
Additionally, our experiments show that both similarity-based selection methods, SemSim and DisSim, enhance performance. However, SemSim performs better on MediaEval, while DisSim excels on Snopes. This indicates the need for further exploration to develop a unified solution capable of handling different data distributions. 
Although experimental results show that using a simple extension on the frozen CLIP model help select useful synthetic data for improving MMD, relying on the CLIP model without fine-tuning for measuring semantic similarity introduces issues like potential bias. This suggests the need for further exploration on how to obtain better multimodal features for improving detection performance.
Lastly, although all data selection methods consistently improve the performance of the base MLLM on the Snopes dataset, they still underperform compared to GPT-4V. This suggests further study on strategy of constructing synthetic datasets for different types of multimodal misinformation.

\section{Ethics Statement}
Our research focuses on detecting real-world multimodal misinformation using synthetic datasets generated by AI technologies, offering a solution to address the scarcity of real-world fact-checking data and enhances the effectiveness of misinformation detection.
Our method is intended for research purposes. To ensure responsible use and prevent potential misuse, we emphasize the necessity of human oversight during utilization to avoid unintended consequences.

All datasets used in our experiments, both synthetic and real-world, are obtained from publicly available sources commonly used in multimodal misinformation detection research. The licenses for public datasets are listed in Appendix~\ref{licences}. There are some image-text pairs in the used  datasets include misleading content that may be disturbing to certain celebrity identities.

\bibliography{anthology,custom}

\appendix

\section{Multimodal Misinformation Data} \label{sec:appendix-dataset}
\subsection{Common Categories} \label{sec:appendix-dataset-cc}
There is no standard categorizations for multimodal information. We categorize them into two types based on whether the image content are falsely altered or counterfeited. Next, we provide a more formulated and detailed introduction. Let $\mathcal{D}= \{(x_i,y_i)\}_{i=1}^{|\mathcal{D}|}$ be a multimodal misinformation dataset with $|\mathcal{D}|$ instances, where each instance consists of the input $x_i$ and the ground-truth label $y_i \in \{0,1\}$. We denote $x_i=(\mathbf{m}_i,\mathbf{t}_i)$ which is a pair of information containing an image $\mathbf{m}_i$ and a corresponding claim text $\mathbf{t}_i$, and $y_i$ indicates the veracity of $x_i$ is true if $y_i=0$ or false otherwise. 

\textbf{Out-of-Context (OOC) Misuse} occurs when the textual claim misrepresents the original context or intent of the image, conveying misleading information. Obtaining OOC image-text pairs by pairing a textual claim with an image taken out-of-context is a straightforward and effective method~\citep{luo-etal-2021-newsclippings,akhtar-etal-2023-multimodal}. Specifically, a mismatched pair $(\mathbf{m}_i,\mathbf{t}_j)$ is created by using two pristine instances $x_i$ and $x_j$ ($j\neq i$). An alternative method involves manipulating the textual claim~\citep{MEIR,TempNews}, by editing the original textual claim $\mathbf{t}_i$ to produce $\tilde{\mathbf{t}}_i$ that is inconsistent with $\mathbf{m}_i$, which may include replacing named entities, altering sentiments or stances, selectively quoting, and other techniques aimed at distorting its meaning to conform to a false narrative.
In this way, a genuine image is accompanied with inconsistent text, conveying misleading information as the text may falsely describe the image's origin, context, and meaning.
Such OOC pairs not only deceive human but also pose significant challenges for automatic misnformation detection models~\cite{luo-etal-2021-newsclippings}.

\textbf{Content Manipulation} involves intentionally altering image and text modality with the aim of deceiving or misleading.
As it can be difficult to consistently obtain real images to support non-factual claim, manipulating image and text content to generate misinformation becomes an alternative approach~\cite{mm-mis-detect}.
Image manipulation refers to altering an image $\mathbf{m}_i$ to obtain a fake image $\tilde{\mathbf{m}}_i$ by modifying the elements within it using techniques such as copying and pasting specific regions, splicing images, removing content, and face swap~\citep{DGM,jia2023autosplice}. Image and text manipulation techniques can be combined to obtain $(\tilde{\mathbf{m}}_i,\tilde{\mathbf{t}}_i)$, ensuring that both visual and textual elements complement each other, thereby strengthening the deceptive information.

\subsection{Training and Evaluation Datasets} \label{sec:appendix-dataset-te}
\subsubsection{Training Datasets}

\textbf{NewsCLIPpings}~\cite{luo-etal-2021-newsclippings} is an automatically generated OOC multi-modal dataset that contains both pristine and falsified instances. It is derived from the VisualNews corpus~\cite{liu-etal-2021-visual}, which contains image-caption pairs from popular news agencies. A falsified instance is an unmanipulated but mismatched image-caption pair, constructing by pairing an image with a caption from an inconsistent context. Given a pristine instance $(\mathbf{m}_i,\mathbf{t}_i)$ as a query, another instance $(\mathbf{m}_j,\mathbf{t}_j)$ is retrieved from the news corpus to form the falsified instance $(\mathbf{m}_j,\mathbf{t}_i)$ ($j \neq i$). The mismatching process is based on the similarity between $(\mathbf{m}_i,\mathbf{t}_i)$ and $(\mathbf{m}_j,\mathbf{t}_j)$, utilizing the CLIP semantic embeddings~\cite{CLIP}, SBERT-WK text embedding~\cite{SBERT-WK}, and scene embeddings~\cite{Places365}. Such image-text mismatch procedure automatically generates around 988k unique synthetic instances. The falsified instances could portray subjects in an image as different entities, depict specific individuals in a misleading context, and mislabel the event depicted within a particular scene. 

$\textbf{DGM}^4$~\cite{DGM} is a multimodal misinformation dataset, where synthetic instances are automatically generated by applying image and text manipulation techniques to pristine instances sourced from the VisualNews corpus. Specifically, for a pristine instance $(\mathbf{m}_i,\mathbf{t}_i)$, a manipulated image $\tilde{\mathbf{m}}_i$ is generated by employing (1) \textit{face swap manipulation} via two representative face swap methods SimSwap~\cite{SimSwap} and InfoSwap~\cite{InfoSwap}, and (2) \textit{face attribute manipulation}, which involves editing the emotion of the main character’s face while preserving the identity using GAN-based methods, HFGI~\cite{HFGI} and StyleCLIP~\cite{StyleCLIP}. 
For text manipulation, $\tilde{\mathbf{t}}_i$ is generated using (1) Sentence-BERT~\cite{reimers-gurevych-2019-sentence} to obtain the text embeddings of $\mathbf{t}_i$ and retrieve a different caption $\mathbf{t}_{j, j \neq i}$ that contains the same person entity as $\mathbf{t}_i$ but has low cosine similarity to the embeddings of $\mathbf{t}_i$, and (2) RoBERTa~\cite{RoBERTa} to classify the sentiment of the $\mathbf{t}_i$ and then replace all sentiment words with the opposite sentiment text to get $\tilde{\mathbf{t}}_i$. After applying these manipulation methods, two manipulated images and two manipulated text captions are generated, along with the original image and caption, resulting in a total of 8 synthetic instances consisting of OOC and manipulation instances for each pristine instance, resulting in a total of 230k synthetic instances.

\textbf{Autosplice}~\cite{jia2023autosplice} is a manipulated image dataset that utilizes a language-image model DALLE-2~\cite{delle2}, guided by textual prompts to edit images. Unlike DGM$^4$, where image and text manipulation are initially employed separately and then combined, Autosplice alters the image based on modified text, enabling a more direct and integrated manipulation process. Specifically, given a pristine image-caption pair $(\mathbf{m}_i,\mathbf{t}_i)$ from the VisualNews corpus, 
an object detection model, Detic~\cite{Detic}, is utilized to extract a list of object regions in $\mathbf{m}_i$, while a text parsing tool, spaCy\footnote{\url{https://github.com/explosion/spaCy}}, is employed to segment text terms in $\mathbf{t}_i$. Subsequently, human annotators select an object region and its corresponding text term, and then substitute this text term with a target generation term that is similar but inconsistent with $(\mathbf{m}_i,\mathbf{t}_i)$.
The modified caption $\tilde{\mathbf{t}}_i$ and image $\mathbf{m}_i$ with the selected object region masked are constructed as inputs for the DALLE-2 model. The model performs local image editing on the masked region and generates the falsified image $\tilde{\mathbf{m}}_i$ based on the text prompt $\tilde{\mathbf{t}}_i$. To facilitate our task, we pair each modified image $\tilde{\mathbf{m}}_i$ with its altered caption $\tilde{\mathbf{t}}_i$, and label it with $y_i = 1$ to indicate false veracity. The pristine instance $(\mathbf{m}_i,\mathbf{t}_i)$ is labeled as $y_i = 0$ to denote truthfulness.

\subsection{Evaluation Datasets}

\textbf{MediaEval}~\cite{mediaeval16} collects a set of tweets associated images/videos with manually verified veracity (i.e., fake or real), which were spread around 17 widely attention-grabbing events such as the November 2015 Paris Attacks\footnote{\url{https://en.wikipedia.org/wiki/November_2015_Paris_attacks}}.  
The number of instances for each event is small, and the classes are imbalanced. Some of these events were hoaxes, hence all instances related to them are fake. Also, there are several events that only contain real instances. For evaluation purposes, we utilize the image-text instances in the test set consisting of 702 instances, in which 292 of them are real and 410 are fake, and remove any replicated instances.

\textbf{Snopes} contains image-related claims from the popular fact-checking website Snopes\footnote{\url{Snopes.com}}. It is initially based on the Fauxtography dataset~\cite{zlatkova-etal-2019-fact}, which consists of image-claim pairs labeled as either True or False on Snopes and image-caption pairs from news agencies.
However, using such news data in evaluation is inappropriate, as it may leak ground-truth labels given that synthetic data is generated using news data. Moreover, news data from trusted news agencies do not require fact-checking. Therefore, we filter the news instances and augment the dataset with more True-labeled instances by combining the fact-checked true image-related claims from the MOCHEG~\cite{end2end-mm} dataset, designed to provide multimodal evidence for claims on Snopes, and additional image-related claims we collected directly from the Snopes website. These image-claim pairs are labeled as either True or False by fact-checkers on Snopes, consisting of 756 instances where which 376 of them are true and 380 are false.

\section{Experiments} \label{sec:appendix-exp}
\subsection{Experimental Settings} \label{sec:appendix-exp-settings}

\subsubsection{Base MLLMs}\label{sec:appendix-exp-mllm}
LLaVA-NeXT-13B~\cite{liu2024llavanext}, an improved version of LLaVA-1.5~\cite{liu2023improved} and LLaVA~\cite{llava}, is an end-to-end trained MLLM that connects the pre-trained CLIP vision encoder ViT-L/14~\cite{ViT-L/14} with an LLM Vicuna-13B~\cite{vicuna2023} using a two-layer MLP as the projection layer. It undergoes a two-stage instruction-tuning process to align two modalities for improving visual reasoning capabilities: (1) pre-training the projection layer with image-text pairs; (2) updating the weights of both the projection layer and LLM using multimodal instruction-tuning data generated by GPT-4~\cite{gpt4}.
LLaVA-NeXT-7B is a smaller version based on an LLM Vicuna-7B.

\textbf{mPLUG-Owl2}~\cite{mPLUG-Owl2} is a versatile MLLM featuring a modularized network design and comprising a vision encoder, a visual abstractor, a text embedding layer, and a language decoder. In lieu of directly aligning the visual features with textual features, mPLUG-Owl2 integrates a modality-adaptive module within the language decoder. This module takes multimodal inputs, utilizing different parameters to project various modalities into a shared semantic space while preserving modality-specific features. The training process involves two stages: (1) pre-training the visual encoder, visual abstractor and newly added parameters of the modality-adaptive module within the language decoder using image-text pairs; (2) instruction-tuning the entire model on unimodal and multimodal instruction data.

\subsubsection{Default Settings}\label{sec:appendix-exp-default}
We utilize the off-the-shelf CLIP model (ViT-L/14)~\cite{ViT-L/14} for feature extraction. We use the original code and pre-trained checkpoint of LLaVA\footnote{\url{https://github.com/haotian-liu/LLaVA}}, with a vision encoder CLIP model (ViT-L/14) and an LLM Vicuna-13B~\cite{vicuna2023}. We fine-tune the LLaVA model using the parameter-efficient fine-tuning method LoRA~\cite{hu2021lora}, with rank set to $128$, $\alpha$ value to $256$, and learned LoRA matrices for both the projector and the LLM. We set training epochs as $3$, batch size as $16$, and learning rate as \num{2e-5} with cosine decay. Following the LLaVA model~\cite{llava}, we convert each training instance into instruction-following data using its specified template. 
We use the API service of GPT-4V from OpenAI\footnote{gpt-4-turbo: \url{https://openai.com/index/gpt-4/}}. During inference, we prompt models using the same prompt template to ensure a fair comparison. 

Given the large scale of the synthetic datasets and our computational constraints, we construct $\mathcal{D}_s$ by randomly sampling 6k instances from the complete training dataset of NewsCLIPings and $\text{DGM}^4$, and 3k instances from the Autosplice dataset since it is smaller, resulting in a total of 15k instances. For a robust evaluation, we repeat the sampling process using three different random seeds, resulting in three distinct $\mathcal{D}_s$. Each data selection method selects 750 instances from $\mathcal{D}_s$ to construct $\mathcal{D}_v$ for fine-tuning, and we ensure the class distribution of selected set is balanced. The unlabeled validation set $\mathcal{D}_u$ contains 5\% instances of each test set, which amounts to $37$ instances for MediaEval and $35$ instances for Snopes and Snopes (O+). 

We conduct experiments three times, each with a different training set, and report the mean macro-F1 and standard deviation of each metric across these three runs in all experiments. The seeds and training dataset are kept the same across different models. All the experiments use a server with 8 NVIDIA Tesla-V100 32GB GPUs.

\section{Visualization of Examples } \label{appendix: vis}
\begin{figure*}[t!]
\centering
\includegraphics[width=0.8\linewidth]{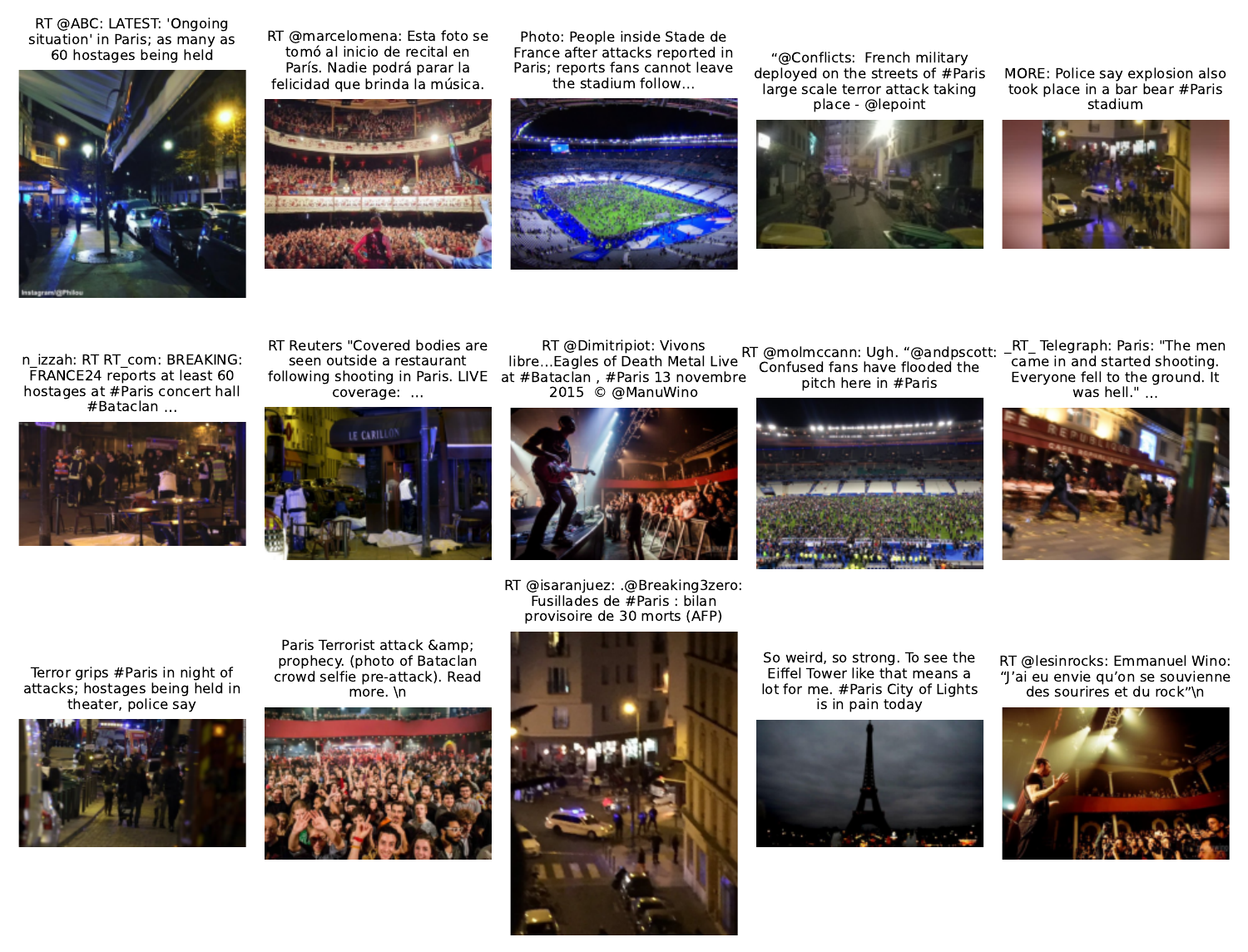}
\caption{Visualization of real-world instances of 2015 Paris Attack event.}
\label{fig:appendix_real_paris}
\end{figure*}

\begin{figure*}[t]
\centering
\includegraphics[width=0.8\linewidth]{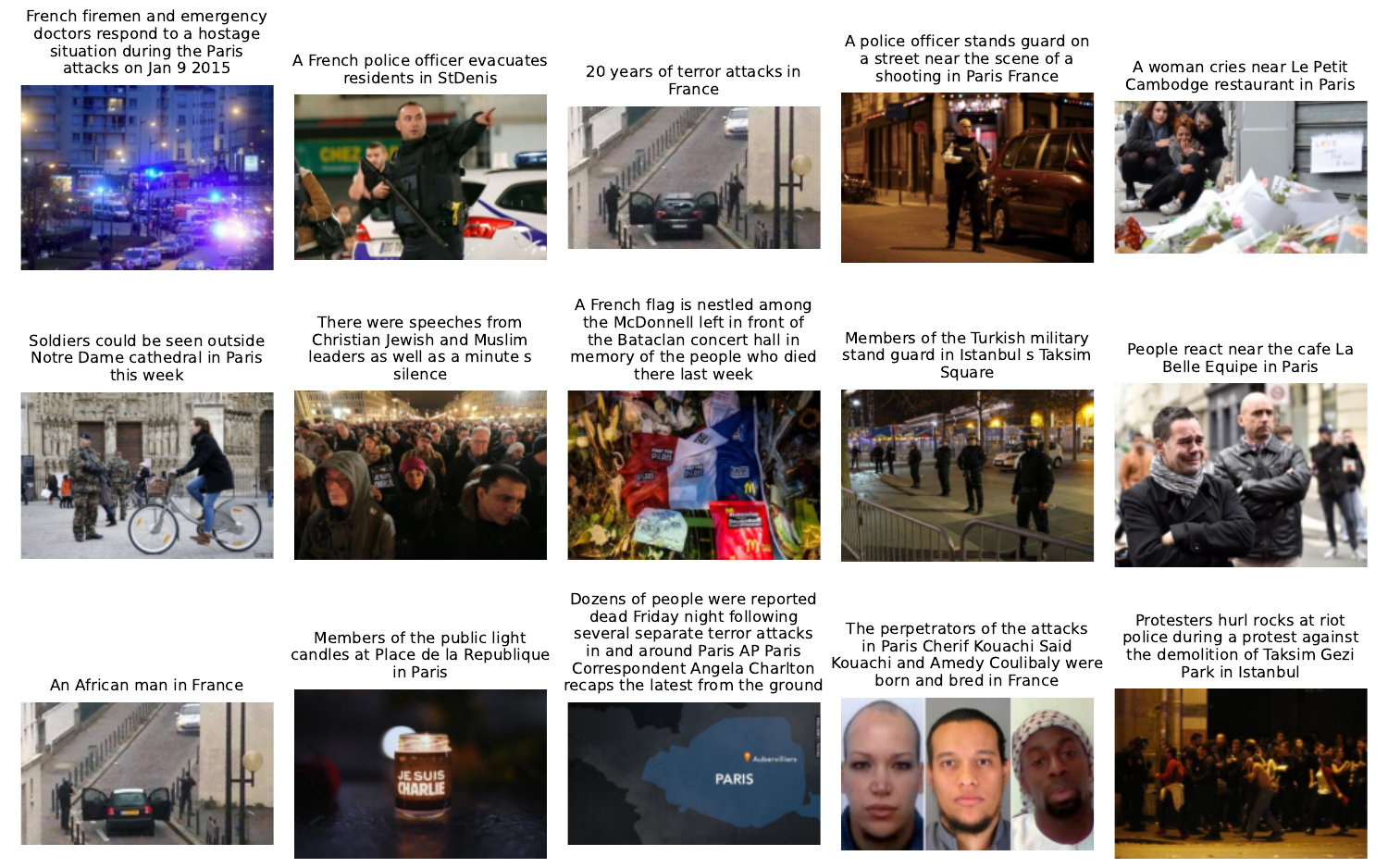}
\caption{Visualization of synthetic instances with high similarity to real-world data.}
\label{fig:appendix_high_paris}
\end{figure*}

\begin{figure*}[t]
\centering
\includegraphics[width=0.8\linewidth]{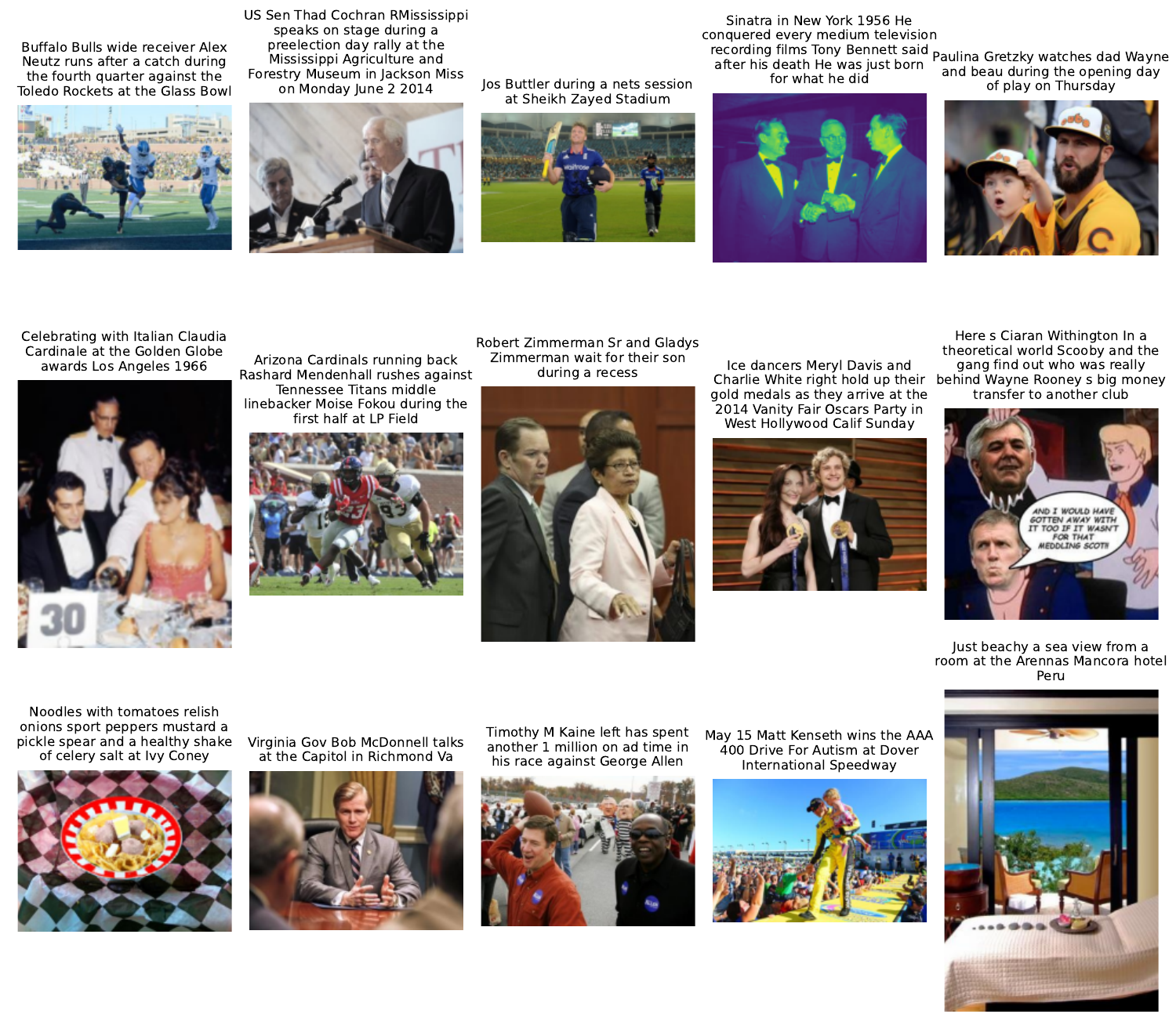}
\caption{Visualization of synthetic instances with low similarity to real-world data.}
\label{fig:appendix_low_paris}
\end{figure*}

We visualize a set of real-world instances from the 2015 Paris Attack event in Figure \ref{fig:appendix_real_paris}. In Figure \ref{fig:appendix_high_paris}, we present a set of synthetic instances with high cosine similarity to these real-world instances, while Figure \ref{fig:appendix_low_paris} shows synthetic instances with low cosine similarity. We observe in Figure \ref{fig:appendix_real_paris} that these instances contain relevant elements such as police, French and Paris, while the instances in \ref{fig:appendix_low_paris} are irrelevant.
These examples illustrate that synthetic datasets not only contain useful instances for multimodal misinformation detection on real-world datasets, but also have irrelevant instances, indicating the gap between synthetic and real-world data.

\section{Licenses of Datasets} \label{licences}
\begin{itemize}[leftmargin=*]

    \item NewsCLIPpings: Unspecified
     \item  $\text{DGM}^4$: Apache-2.0
    \item AutoSplice: Only for academic research
    \item MediaEval: Apache-2.0
    \item Fauxtography: MIT License
    \item MOCHEG: CC BY 4.0

\end{itemize}

\end{document}